\def\year{2018}\relax
 \documentclass[letterpaper]{article} 
\usepackage{aaai18}  
\usepackage{times}  
\usepackage{helvet}  
\usepackage{courier}  
\usepackage{url}  
\usepackage{graphicx}  
\usepackage{longtable}
\usepackage{booktabs}
\usepackage{multirow}
\usepackage{hhline}
\usepackage{pifont}
\usepackage{paralist}
\usepackage{enumitem}
\usepackage{algorithm}
\usepackage{algpseudocode}
\usepackage{amsmath}
\usepackage{amssymb}

\begin{document}
%
\title{Denoising Dictionary Learning Against Adversarial Perturbations}
\author{John Mitro, Derek Bridge, Steven Prestwich \\
  \{j.mitro, d.bridge, s.prestwich\}@insight-centre.org \\
  Insight Centre for Data Analytics \\
  Department of Computer Science \\
  University College Cork, Ireland
}
\maketitle

\begin{abstract}
We propose denoising dictionary learning (DDL), a simple yet effective
technique as a protection measure against adversarial
perturbations. We examined denoising dictionary learning on MNIST and
CIFAR10 perturbed under two different perturbation techniques, fast
gradient sign (FGSM) and jacobian saliency maps (JSMA). We evaluated
it against five different deep neural networks (DNN) representing the
building blocks of most recent architectures indicating a successive
progression of model complexity of each other. We show that each model
tends to capture different representations based on their
architecture. For each model we recorded its accuracy both on the
perturbed test data previously misclassified with high confidence and
on the denoised one after the reconstruction using dictionary
learning. The reconstruction quality of each data point is assessed by
means of PSNR (Peak Signal to Noise Ratio) and Structure Similarity
Index (SSI). We show that after applying (DDL) the reconstruction of
the original data point from a noisy sample results in a correct
prediction with high confidence.
\end{abstract}

\section{Introduction}
The observation of adversarial perturbations indicated in the seminal
work by~\cite{szegedy2013:intriguing-properties-of-}, is formulated as
an attack against DNN
models~\cite{papernot2016:towards-the-science-of-se}. It describes a
mechanism for devising noise taking into account the models'
output. This misbehavior is described as a high confidence
involuntary misclasification on the models' part, which could
potentially undermine the security of environments where these models
are deployed. We examined five DNN models,
\begin{inparaenum}[(i)]
  \item multilayer perceptron (MLP),
  \item convolutional neural network (CNN),
  \item auto-encoder (AE),
  \item residual network (RNet),
  \item hierarchical recurrent neural network (HRNN),
\end{inparaenum}
of varying complexity and topology, in order to identify how they
respond under adversarial noise.

\begin{figure}
\includegraphics[width=\columnwidth]{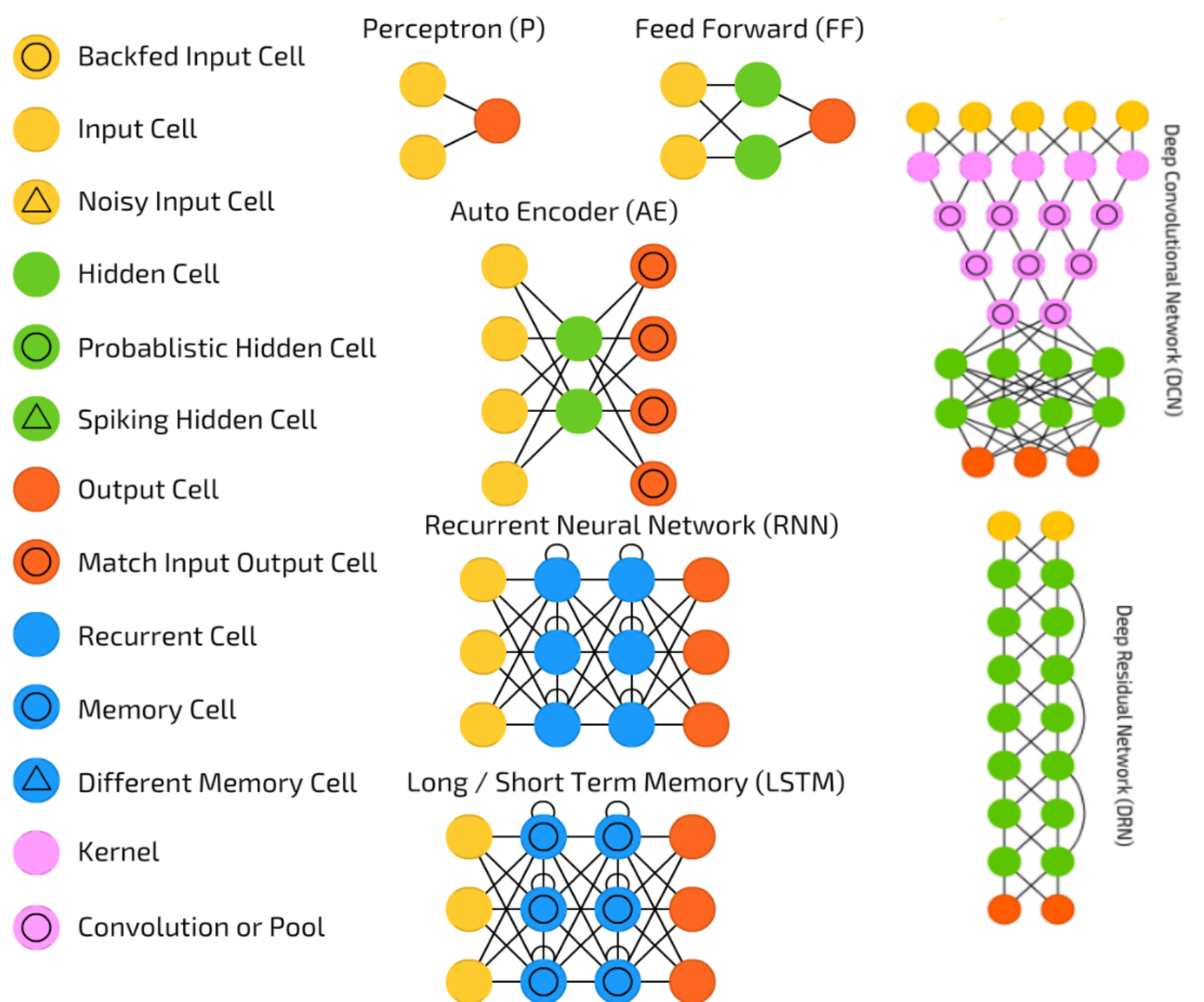}
\caption{Deep neural network models tested against adversarial
perturbations~\cite{veen_neural_2016}.}
\label{fig:architectures}
\end{figure}

   We investigate and visualize which parts of the data maximize the
output of the model and verify whether those areas are contaminated under
these perturbation attacks. Furthermore, we propose denoising
dictionary learning as a measure of protection against adversarial
perturbations. It exhibits desirable properties such as
robustness~\cite{gregor2010:learning-fast-approximati},
\cite{kavukcuoglu2010:fast-inference-in-sparse-} and
flexibility~\cite{szlam2012:fast-approximations-to-st},
\cite{thom2015:efficient-dictionary-lear} since it can be incorporated
in any supervised learning
algorithm~\cite{mairal2008:supervised-dictionary-lea}. Furthermore, it
can operate without the presence of noisy samples which indicates a
common use case in adversarial attacks where only the test data are
perturbed and presented to the model. In comparison
to~\cite{wang2016:using-non-invertible-data} denoising dictionary
learning has the ability to recover signals from heavily noised
samples, moreover, it does not render the data unreadable compared
to~\cite{wang2016:using-non-invertible-data} which performs a non
invertible transformation. After the transformation the data are
deemed noninterpretable therefore it would require the storage of
both datasets, prior to and after the transformation, for
interpretability of the results. This not only increases the storage
space but is also computationally inefficient in real world scenarios.
Current DNN models can be easily exploited by a varying number of
different attacks~\cite{papernot2016:crafting-adversarial-inpu},
~\cite{grosse2016:adversarial-perturbations},
~\cite{moosavi-dezfooli2015:deepfool:-a-simple-and-ac},
~\cite{carlini2016:towards-evaluating-the-ro},
~\cite{robinson2015:confus-deep-convol-networ-relab},
~\cite{narodytska2016:simple-black-box-adversar},
~\cite{sethi17:data_driven_explor_attac_black}, some of them require
knowledge of the intrinsic structure of the
model~\cite{goodfellow2014:explaining-and-harnessing} while others can
operate without any prior knowledge of the loss function of the
model~\cite{papernot2016:practical-black-box-attac}. These
perturbations can be categorized in three categories.
\begin{inparaenum}
  \item Model specific perturbations,
  \item White-box perturbations,
  \item Black-box perturbations.
\end{inparaenum}
  It is important to notice that there might be an overlap between
different perturbation techniques. For the experiments conducted in this
study two adversarial perturbation attacks have been utilized. The
first one called fast gradient
sign~\cite{goodfellow2014:explaining-and-harnessing}, described as the
gradient of the loss of the model multiplied by a scalar. The second
one is called jacobian saliency
map~\cite{papernot2015:the-limitations-of-deep-l} and is exploiting
the forward derivative rather than the cost function emitting
information about the learned behavior of the model. The
differentiation is applied with respect to the input features rather
than the network parameters. Instead of propagating gradients
backwards it propagates them forward which permits to find those
pixels in the input image that lead to significant changes in the
network output.  Our contributions in this study are two fold. First,
we provide an intuitive explanation of the main building operations or
components of DNN, how they operate and how they can be utilized to
build more complex models. In addition, we describe which components
remain unchanged and how they might cause adversarial
perturbations. Finally, we provide a defense mechanism against
adversarial perturbations based on sparse dictionary learning to
alleviate the problem.

\section{Theoretical Background}
In the following Section~\ref{sec:NN} we provide the necessary
information required to understand how DNN operate given their basic
building blocks. Based on that information we show how to construct
more complex models and identify the main problematic components
leading to adversarial perturbations. The ``\textit{Adversarial
Perturbations}'' Section~\ref{sec:perturbations} describes two well
known adversarial perturbations utilized during the experiments
indicating which components are harnessed in order to devise the
perturbations. Finally, the ``\textit{Dictionary Learning}''
Section~\ref{sec:dictionary} describes in detail how the proposed
defense mechanism is devised, constructed and how it operates to
reverse the effect of the perturbations.

\subsection{Deep Neural Network Components}
\label{sec:NN}
\begin{figure}[!htbp]
\centering
\includegraphics[width=6cm]{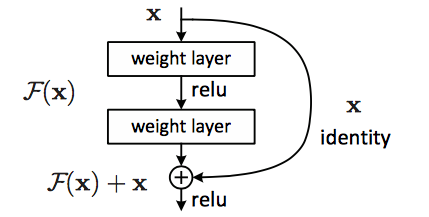}
\caption{Residual network building block.}
\label{fig:resnet}
\end{figure}

Each of the models illustrated in Figure~\ref{fig:architectures} is
color coded in order to denote the different building components. Each
block describes a different operation. Here we provide a formal
representation of each models' structure starting from a model as
simple as a mulitlayer perceptron (MLP) up until to an LSTM unit. An
(MLP) with one hidden layer can be described as a function $f:
\mathbb{R}^{d}\;\rightarrow\;\mathbb{R}^{p}$ where $d$ is the size of
the input vector $\vec{x}$ and $p$ is the size of the output vector
$\vec{y} = f(\vec{x})$. Using matrix notation we formulate it as:
\begin{equation}
\label{eqn:mlp}
  \vec{y} =
  \alpha(\mathbf{W}^{(\ell)}\; \alpha\;(\mathbf{W}^{(\ell-1)}\;\vec{x}\; +\; \vec{b}^{(\ell-1)})\; +\;
  \vec{b}^{(\ell)})
\end{equation}
where $\alpha(\cdot)$ is the activation function. Setting $h(\vec{x})
= \alpha(\mathbf{W}^{(\ell-1)}\;\vec{x}\; +\; \vec{b}^{(\ell-1)})$ we can rewrite the above
equation as:
\begin{equation}
\label{eqn:mlp_layer}
\vec{y} = \alpha(\mathbf{W}^{(\ell)}\; h(\vec{x}) + \vec{b}^{(\ell)})
\end{equation}
On the same notion we can extend Equation~\ref{eqn:mlp} to accommodate
for convolutional operations. Recall that a convolution is defined as:
\begin{equation}
\label{eqn:continuous_convolution}
f\;\ast\; g(x) = \int_{\Omega} f(y)\; g(x - y)\; dy
\end{equation}
For the discrete case of a 1D signal the formulation is:
\begin{equation}
  \begin{split}
    y[n] &= f[n]\;\ast\; g[n] \\
         &= \sum_{u=-\infty}^{\infty} f[u]\;g[n - u] \\
         &= \sum_{u=-\infty}^{\infty}f[n - u]g[u]
  \end{split}
\end{equation}
This can be extended to 2D as follows:
\begin{equation}
\label{eqn:2d_conv}
  \begin{split}
    y[m, n] &= f[m, n]\ast g[m, n] \\
            &= \sum_{u=-\infty}^{\infty}\sum_{v=-\infty}^{\infty} f[u, v]\;g[m-u, n-v]
  \end{split}
\end{equation}
Replacing Equation~\ref{eqn:2d_conv} with
Equation~\ref{eqn:mlp_layer} we get the representation for the
$\ell^{\text{(th)}}$ convolutional hidden layer:
\begin{equation}
  \vec{h}_{ij}^{(\ell)} = \alpha((\mathbf{W}^{(\ell)} \ast \vec{x})_{ij} + \vec{b}_{j}^{(\ell)})
\end{equation}
We can easily extend the formulation of a convolutional layer to define
auto-encoders which could be described as a succession of hidden
layers mapped first into a latent space $\vec{z}$ instead of the
original space $\vec{x}$ and finally with an inverse transformation
restored to the original space:
\begin{equation}
  \begin{split}
    \vec{z} &= \alpha(\mathbf{W}^{(\ell)}\vec{x} + \vec{b}^{(\ell)}) \\
    \tilde{\vec{x}} &= \alpha({\mathbf{W}^{(\ell)}}^{T} \vec{z} + \vec{b}^{(\ell)})
  \end{split}
\end{equation}
Even residual
networks~\cite{he2015:deep-resid-learn-image-recog} and their
residual building block shown in Figure~\ref{fig:resnet}.

can be reconstructed from Equation~\ref{eqn:mlp_layer} as follows:
\begin{equation}
\label{eqn:residual}
  \vec{h}^{(\ell)} =
  \alpha((\mathbf{W}^{(\ell)}\; \alpha\;(\mathbf{W}^{(\ell-1)}\;\vec{x}\; +\; \vec{b}^{(\ell-1)})\; +\;
  \vec{b}^{(\ell)}) + \vec{x})
\end{equation}
Notice that the bias term $\vec{b}$ is optional and in most recent residual
network architectures is omitted.

Finally, recurrent neural networks as a building block introduce two
new concepts. The concept of time and memory. Time could be easily
described as a feed forward network unfolded across the time
axis. Memory permits the network to share states across the different
hidden layers.
\begin{equation}
  \begin{split}
    \vec{h}^{(t)} &= \alpha(\mathbf{W}^{(xh)}\vec{x}^{(t)} +
    \mathbf{W}^{(hh)}\vec{h}^{t-1} + \vec{b}^{(h)}) \\
    \vec{h}^{(t)} &= \alpha(\mathbf{W}\vec{x}^{(t)} + \mathbf{U}\vec{h}^{(t-1)})
  \end{split}
\end{equation}
The hidden state $\vec{h}^{(t)}$ at time step $t$ represents a
function of the input $\vec{x}^{(t)}$ modified by a weight matrix
$\mathbf{W}$ added to the
hidden state of the previous time step $\vec{h}^{(t-1)}$, multiplied by
its own hidden to hidden state matrix $\mathbf{U}$, otherwise known as
a transition matrix, and similar to a Markov chain. The weight matrices
are filters that determine the amount of importance to accord to both
the present input and the past hidden state.  LSTMs, first introduced
by~\cite{hochreiter1997:long-short-term-memory}, can be described as a
collection of gates with additional constraints. An LSTM layer is
described as:
\begin{align}
  \begin{split}
    \vec{i}^{(t)} &= \alpha(\mathbf{W}^{(ix)}\vec{x}^{(t)} +
    \mathbf{W}^{(ih)}\vec{h}^{(t)} + \vec{b}^{(i)}) \\
    \vec{g}^{(t)} &= \alpha(\mathbf{W}^{(ix)}\vec{x}^{(t)} + \mathbf{W}^{(ih)}\vec{h}^{(t-1)} +
    \vec{b}^{(i)}) \\
    \vec{f}^{(t)} &= \alpha(\mathbf{W}^{(fx)}\vec{x}^{(t)}
    + \mathbf{W}^{(fh)}\vec{h}^{(t-1)} + \vec{b}^{(f)}) \\
    \vec{o}^{(t)} &= \alpha(\mathbf{W}^{(ox)}\vec{x}^{(t)} +
    \mathbf{W}^{(oh)}\vec{h}^{(t-1)} + \vec{b}^{(o)}) \\
    \vec{c}^{(t)} &= \vec{i}^{(i)}\odot \vec{g}^{(t)} + \vec{c}^{(t-1)}\odot
    \vec{f}^{(t)} \\
    \vec{h}^{(t)} &= \alpha(\vec{c}^{(t)}) \odot\vec{o}^{(t)}.
  \end{split}
\end{align}

In this section we wanted to demonstrate how complex DNN models
can be constructed starting from MLP up to LSTMs.
Even though the different components might
require some changes when transitioning from one DNN model to the
other, usually three of them remain consistent. Those refer to
\begin{inparaenum}[(a)]
  \item the loss function, which is the categorical cross entropy in
  Equation~\ref{eqn:loss} for all five models,
  \item the activation function ReLU,
  \item the optimization process resulting in a variant of gradient
descent~\cite{kingma2014:adam-optimizer}.
\end{inparaenum}

We believe that is this combination that is responsible for the
phenomenon of adversarial perturbations. Examining closely each
building block we realize that DNN models resemble linear models
since most of the components either are a result of, or describe a
linear transformation. This would explain the high confidence in
misclassified examples since linear models also have the tendency to
extrapolate to unseen data points with high confidence. Unfortunately,
the loss function does not help either, in this situation, since
eventually any differential loss has the potential to emit
information on which parts of the input should be altered accordingly
in order to maximize a particular output target. This information can
be exploited by an adversary. Finally, any optimization process
requiring small steps towards the direction of the local minimum such
as gradient descent or variants will result in a process where small
changes gradually lead to overall bigger effects. It is these small
changes that the adversaries exploit in order to devise the
perturbations.

\subsection{Adversarial Perturbations}
\label{sec:perturbations}
In this section we provide a short description and formulation of the
adversarial perturbations utilized during the experiments. First, we
present (FGSM) which is the gradient of the loss
$\nabla_{\vec{x}}\mathbb{L}(\vec{x}, \vec{y})$ with respect to the
input $\vec{x}$ multiplied by a constant $\epsilon$ defined as:
\begin{align}
  \label{eqn:loss}
 \begin{split}
    \epsilon \nabla_{\vec{x}} \mathbb{L}(\vec{x}, \vec{y}) &=
    -\frac{1}{N}\sum_{n=1}^{N}\nabla_{\vec{x}}\mathbb{L}
    y^{(n)}\ln\alpha(\vec{x}^{(n)}) \\
    & + (1-y^{(n)})\ln(1-\alpha(\vec{x}^{(n)}))
 \end{split}
\end{align}
where $\alpha(\vec{x})$ describes the output of the neural network
given input $\vec{x}$. The final perturbed sample for an input
$\vec{x}$ is $\vec{x} + \epsilon\nabla_{\vec{x}}\mathbb{L}(\vec{x},
\vec{y})$. Second, we present (JSMA) in the following three steps.
\begin{inparaenum}[(i)]
  \item Compute the forward derivative $\nabla_{\vec{x}*}f(\vec{x}*)$.
  \item Construct the saliency map $\Sigma$ based on the derivative.
  \item Modify an input feature $m$ by $\epsilon$.
\end{inparaenum}
The adversary's' objective is to craft an adversarial sample $\vec{x}*
= \vec{x} + \delta_{\vec{x}}$ such that the final output of the
network results in a misclasification $f(\vec{x}*) = \hat{\vec{y}}
\neq \vec{y}$ where $f_{n}$ describes the derivative of one output
neuron. Reformulating it as an optimization problem we have the
following objective
$\min_{\delta_{\vec{x}}}\Vert\delta_{\vec{x}}\Vert$ $\mbox{ s.t. }
f(\vec{x} + \delta_{\vec{x}}) = \hat{\vec{y}}$. The forward derivative
of a DNN for a given sample is essentially the Jacobian learned by the
neural network $\nabla f(\vec{x}) = \left[\frac{\partial
f_{n}(\vec{x})}{\partial x_{m}}\right]_{m=1\ldots,M, n=1,\ldots,N}$,
where $m$ denotes the input feature and $n$ denotes the output of
neuron $n$.

Next compute the saliency map based on the forward
derivative. Saliency maps convey information which pixel intensity
values should be increased in order for a specific target $t$ to be
misclassified by the neural network $t\neq y(\vec{x})$ where
$y(\vec{x})$ denotes the true label assigned to input $\vec{x}$. The
saliency map $\Sigma(\vec{x}, t)$ is defined as:
\begin{align}
  \Sigma(\vec{x}, t)_{m} =
  \begin{cases}
    0 \mbox{  if  } \frac{\partial f_{t}(\vec{x})}{\partial x_{m}} < 0 \mbox{
      or }
    \sum_{n\neq t}\frac{\partial f_{n}(\vec{x})}{\partial x_{m}} \\
    \left(\frac{\partial f_{t}(\vec{x})}{\partial x_{m}}\right) \;\left\vert \sum_{n\neq
        t}\frac{\partial f_{n}(\vec{x})}{\partial x_{m}}\right\vert
  \end{cases}
\end{align}
The first line rejects components with negative target derivative or
positive derivative on all other classes. The second line considers
all other forward derivatives. In summary, high saliency map values
denote features that will either increase the target class or decrease
other classes significantly. Increasing those feature values causes
the neural network to misclasify a sample into the target class.

\subsection{Dictionary Learning}
\label{sec:dictionary}

In summary, dictionary learning could be deconstructed into the
following three principles.
\begin{inparaenum}[(i)]
	\item Linear decomposition,
	\item sparse approximation, and
	\item dictionary learning.
\end{inparaenum}
Linear decomposition asks the question whether a signal $\vec{y}$ can
be described as a linear combination of some basis vectors and their
coefficients. Given a signal $\vec{y}\in\mathbb{R}^{N}$ and a matrix
$\mathbf{D}\in\mathbb{R}^{M\times N}$ of $N$ vectors $[d]_{n=1}^{N}$,
the linear decomposition of $\vec{y}$ described by $\mathbf{D}$ is
such that $\vec{y} = \mathbf{D}\vec{s} + \vec{\epsilon} =
\sum_{n=1}^{N} \vec{d}_{n}\vec{s}_{n} + \vec{\epsilon}$ where
$\vec{s}\in\mathbf{R}^{N}$ represent the coefficients and
$\vec{\epsilon}$ is the error. Sparse approximation on the other hand
refers to the ability of $\mathbf{D}$ to reconstruct $\vec{y}$ from a
set of sparse basis vectors. When $\mathbf{D}$ contains more vectors
than samples, i.e., $N > M$, then it is called a dictionary and its
vectors are referred to as atoms. Since $\vec{y} = \mathbf{D}\vec{s} +
\vec{\epsilon}$, has multiple solutions for $\vec{s}$. It is usually
accustomed to introduce different types of constraints, such as, for
instance, sparsity or others, in order to allow the solution to be
regularized. The decomposition of $\vec{y}$ under sparse constraints
is formulated as $\arg\min_{\vec{s}} \Vert\vec{y} -
\mathbf{D}\vec{s}\Vert_{2}^{2}\;\mbox{s.t.}\; \Vert\vec{s}\Vert_{1}
\le T$, where $T$ is a constant. There exist a plethora of algorithms
to deal with this problem, some utilize greedy approaches such as
orthogonal matching pursuit
(OMP)~\cite{pati1993:orthogonal-matching-pursu} or $\ell_{1}$-norm
based optimizations such as basis
pursuit~\cite{chen1998:atomic-decomposition-by-b}, least-angle
regression~\cite{efron2004:least-angle-regression}, iterative
shrinkage thresholding~\cite{daubechies2004:an-iterative-algorithm-fo}
which guarantee convex properties.

OMP proceeds by iteratively selecting the atoms, i.e., the columns in
a dictionary with corresponding nonzero coefficients
$\vec{s}\in\mathbb{R}^{N}$ computed via orthogonal projection of
$\vec{y}$ on $\mathbf{D}$ that best explain the current residue
$\vec{\epsilon} = \vec{y} - \mathbf{D}\vec{y}$. Mainly the
optimization process is a two step approach alternating between OMP
and dictionary learning. As we mentioned the goal of dictionary
learning is to learn the dictionary $\mathbf{D}$ that is most suitable
to sparsely approximate a set of signals as it is shown in
Equation~\ref{eqn:dictionary-learning}. This non-convex problem is
usually solved by alternating between the extraction of the main
atoms, which is referred to as sparse coding or sparse approximation
step, and the actual learning process which is referred to as the
dictionary update step. This optimization scheme reduces the error
criterion iteratively. There are several dictionary learning
algorithms, such as maximum likelihood (ML), method of optimal
directions (MOD), and K- SVD for batch methods, and online dictionary
learning and recursive least squares (RLS) for online methods, which
are less expensive in computational time and memory than batch
methods.

Next we will formulate the problem of dictionary learning from
corrupted samples and demonstrate its ability as a defense mechanism
against adversarial perturbations. Formally, the problem of image
denoising is described as follows:
\begin{equation}
  \vec{y} = \vec{x} + \vec{\epsilon}
\end{equation}
$\vec{y}$ is our measurements, $\vec{x}$ is the original image and
$\vec{\epsilon}$ is the noise. The objective is to recover $\vec{x}$ from our
noisy measurements $\vec{y}$. We can reformulate our problem as an
energy minimization problem also known as maximum a posterior
estimation.
\begin{equation}
  \mathbb{E}[\vec{x}] = \Vert \vec{y} - \vec{x} \Vert_{2}^{2} + \text{Pr}(\vec{x})
\end{equation}
$\Vert \vec{y} - \vec{x} \Vert_{2}^{2}$ refers to the relation to the measurements and
$\text{Pr}(\vec{x})$ is the prior. There are a number of different
classical priors from which we can choose.
\begin{itemize}
  \item Smoothness $\lambda\Vert\mathcal{L}\vec{x}\Vert_{2}^{2}$
  \item Total variation $\lambda\Vert\nabla \vec{x}\Vert_{1}^{2}$
  \item Wavelet sparsity $\lambda\Vert \mathbf{W}\vec{x}\Vert_{1}$
\end{itemize}
Utilizing sparse representations for image reconstruction we can rewrite
$\text{Pr}(\vec{x}) = \lambda\Vert \vec{s}\Vert_{0}\; \mbox{for}\; \vec{x} \approx
\mathbf{D}\vec{s}$. This could be translated visually to the following
operation's:
 \begin{displaymath}
   \underbrace{
   \begin{pmatrix}
      \\
      y \\
      \\
   \end{pmatrix} }_\text{$\vec{y} \in~\mathbb{R}^{M}$} =
   \underbrace{
   \begin{pmatrix}
     & &  & &  &  &  \\
     \mathbf{d}_{1} & & \mathbf{d}_{2} & & \cdots & & \mathbf{d}_{N} \\
     & &  & &  &  &
   \end{pmatrix} }_\text{$\mathbf{D}\in \mathbb{R}^{M\times N}$}
   \underbrace{
   \begin{pmatrix}
     \mathbf{s}_{1} \\
     \mathbf{s}_{2} \\
     \vdots \\
     \mathbf{s}_{N} \\
   \end{pmatrix}
   }_\text{$\vec{s}\in \mathbb{R}^{N}$, sparse}
 \end{displaymath}

Learning a dictionary of atoms can be viewed as an optimization
equation derived from two parts. The first part refers to the
reconstruction of the original signal and the second part refers to
the sparsity.

\begin{equation}
\label{eqn:dictionary-learning}
  \min_{\vec{s}_{n}, \mathbf{D}\in C} \sum_{n=1}^{N}
\frac{1}{2}\Vert\vec{y_{n}} - \mathbf{D}\vec{s}_{n}\Vert_{2}^{2} + \lambda\phi(\vec{s}_{n})
\end{equation}

Where $\phi(\vec{s}) = \Vert \vec{s}\Vert_{0}$ is the $\ell_{0}$
pseudo-norm and $\phi(\vec{s}) = \Vert \vec{s} \Vert_{1}$ is the
$\ell_{1}$ norm. How the optimization problem works is as follows. We
formulate and solve a matrix factorization problem after we have
extracted all overlapping $8\times 8$ patches of $\vec{y}$. The
factorization problem is formulated as follows:
\begin{equation}
  \begin{split}
  \min_{\vec{s}_{n}, \mathbf{D}\in C} \sum_{n=1}^{N}\underbrace{\frac{1}{2}\Vert \vec{y}_{n} - \mathbf{D}\vec{s}_{n}\Vert_{F}
  ^{2}}_\text{reconstruction} +
\underbrace{\lambda\phi(\vec{s}_{n})}_\text{sparsity} \\
  \frac{1}{2}\Vert\mathbf{Y} - \mathbf{D}\vec{s}_{n}\Vert_{F}
  ^{2} + \lambda\Vert\vec{s}_{n}\Vert_{1} \\
  \mbox{where}\; \lambda\phi(\vec{s}_{n}) = \Vert\vec{s}\Vert_{1} \\
  \mbox{such that}\; \mathbf{Y} = [\vec{y}_{1}, \ldots,
  \vec{y}_{n}]\; \mbox{and}\; \vec{s} = [s_{1}, \ldots, s_{n}].
  \end{split}
\end{equation}
Notice that different constraints adhere on $\mathbf{D}$ and $\vec{s}$
depending on the matrix factorization approach. For instance if PCA is
selected as a solution to the factorization problem then $\mathbf{D}$
should be orthonormal and $\vec{s}^{T}$ orthogonal. Otherwise if non
negative matrix factorization is selected then $\mathbf{D}$ and
$\vec{s}$ should be non negative. The optimization for dictionary
learning is formulated as follows:
\begin{align}
  \begin{split}
    \min_{\mathbf{D}\in C}\ f(\mathbf{D}) &=
    \mathbb{E}_{\vec{x}}[\xi(\vec{y}, \mathbf{D})] \\
    & \approx \lim_{n\rightarrow\infty+}\frac{1}{N}\sum_{n=1}^{N}\xi(\vec{y}_{n}, \mathbf{D}) \\
    \mbox{where}
    \min_{\mathbf{D}\in C}f(\mathbf{D}) &= \min_{\mathbf{D}\in C}
    \frac{1}{N}\sum_{n=1}^{N}\xi(\vec{y}_{n}, \mathbf{D}) \\
    C \triangleq \mathbf{D}\in \mathbb{R}^{M\times N}\; |\; \forall n &=
    1,\ldots,N \;\;\mbox{s.t.}\;\; \Vert\mathbf{d}_{n}\Vert_{2}\; \le\; 1 \\
    \xi(\vec{y}, \mathbf{D}) & \triangleq \min_{\vec{s}\in \mathbb{R}^{N}}\frac{1}{2}\Vert
    \vec{y} - \mathbf{D}\vec{s}\Vert_{2}^{2} + \lambda\Vert \vec{s}\Vert_{1}.
  \end{split}
\end{align}

In the following section we provide the description for OMP and
dictionary learning algorithms utilized in the experiments. Regarding
Algorithm 1, at the current iteration $t$, OMP selects the atom
$\hat{\alpha}$ that produces the strongest decrease in the
residue. This is equivalent to selecting the atom that is most
correlated with the residue. An active set, $Q$ i.e., nonzero, is
formed, which contains all of the selected atoms. In the following
step the residue $\epsilon$ is updated via an orthogonal projection of
$\vec{y}$ on $\mathbf{D}$. Finally, the sparse coefficients of
$\vec{s}$ are also updated according to the active set $Q$.

\begin{algorithm}[ht]
\label{alg:omp}
  \caption{Orthogonal matching pursuit algorithm.}
  \begin{algorithmic}
    \State $\min_{\vec{s}\in\mathbf{R}^{N}}\Vert\vec{y} -
    \mathbf{D}\vec{s}\Vert_{2}^{2}\;\mbox{s.t.}\; \Vert\vec{s}\Vert_{1} \le T$
    \State $Q = \varnothing$
    \For{$t=1$ to $T$}
    \State \mbox{Select the atom that reduces the objective}
    \State $\hat{\alpha}\gets \arg\min_{t \in Q^{c}}\{\min_{\vec{s}}\Vert\vec{y} -
      \mathbf{D}_{Q\cup{t}}\vec{s}\Vert_{2}^{2}\}$
    \State \mbox{Update the active set:} $Q\gets Q\cup\hat{\alpha}$,
    \State \mbox{Update the residual via orthogonal projection:}
    \State $\vec{\epsilon}\gets (\mathbf{I} -
    \mathbf{D}_{Q}{(\mathbf{D}_{Q}^{T}\mathbf{D}_{Q})}^{-1}\mathbf{D}_{Q}^{T})\vec{y}$
    \State \mbox{Update the coefficients:} $\vec{s}_{Q}\gets
    {(\mathbf{D}_{Q}^{T}\mathbf{D}_{Q})}^{-1}\mathbf{D}_{Q}^{T}\vec{y}$
   \EndFor{}
  \end{algorithmic}
\end{algorithm}

As for Algorithm 2 the sparse coding step is usually the one which is
carried out by orthogonal matching pursuit described in Algorithm 1.

\begin{algorithm}[ht]
  \caption{Dictionary optimization algorithm.}
  \begin{algorithmic}
    \Require $\mathbf{D} \in \mathbb{R}^{m\times n}, \; \lambda\in \mathbb{R}$
    \State $\mathbf{A} = 0, \mathbf{B} = 0$
    \For{$t=1$ to $T$}
    \State Draw $\mathbf{y}_{t}$
    \State $\mathbf{s}_t\gets$
    $\arg\min_{\mathbf{s}\in\mathbb{R}^{n}}\frac{1}{2}\Vert\mathbf{y}_{t}
    - \mathbf{D}_{t-1}\mathbf{s}\Vert_{2}^{2} + \lambda\Vert\mathbf{s}\Vert_{1}$,\Comment{Sparse Coding}
    \State $\mathbf{A}_{t}\gets\mathbf{A}_{t-1} +
    \mathbf{s}_{t}\mathbf{s}_{t}^{T}$,
    \State $\mathbf{B}_{t}\gets \mathbf{B_{t-1} + \mathbf{y}_{t}\mathbf{s}_{t}^{T}}$
    \State $\mathbf{D}_{t}\gets\arg\min_{\mathbf{D\in
        C}}\frac{1}{N}\sum_{n=1}^{N}(\frac{1}{2}\Vert \mathbf{y}_{n} -
    \mathbf{D}\mathbf{s}_{n}\Vert_{2}^{2} + \lambda\Vert\mathbf{s}_{n}\Vert_{1})$
   \EndFor{}
  \end{algorithmic}
\end{algorithm}

\section{Methodology}
\label{sec:methodology}
In this study we trained five different models for 100 epochs, from
multilayer perceptrons to hierarchical LSTMs with a batch size of 32,
on two different datasets, MNIST and CIFAR10 whose distributions are
presented below.

\begin{figure}[ht]
  \centering
  \includegraphics[width=8cm]{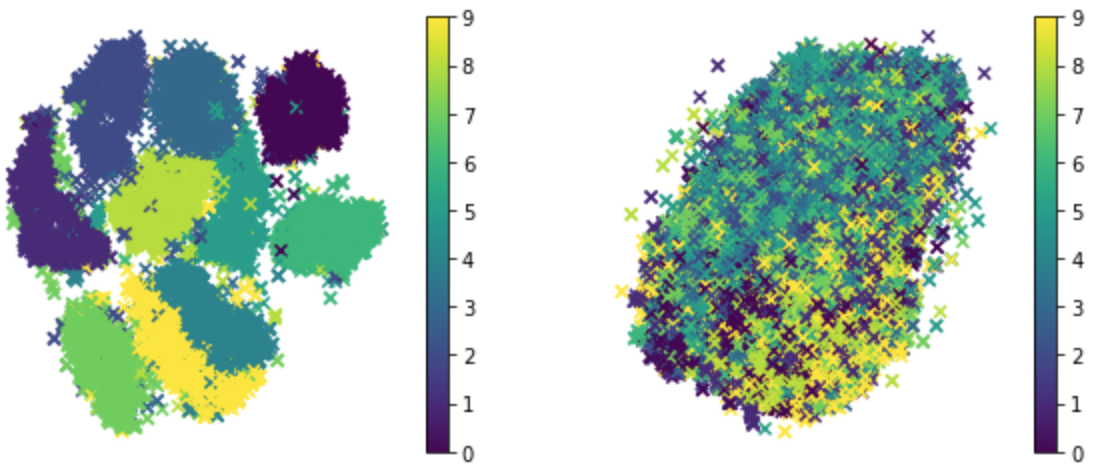}
\caption{T-SNE embedding of MNIST (left) and CIFAR10 (right).}
\label{fig:tsne-embedding}
\end{figure}

\begin{figure}[ht]
\begin{tabular}{cc}
  \includegraphics[width=3.79cm]{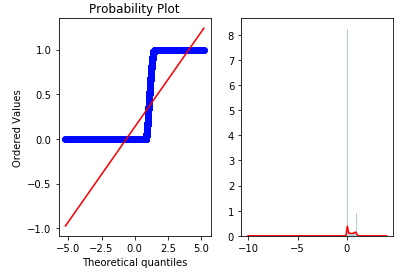} &
  \includegraphics[width=3.79cm]{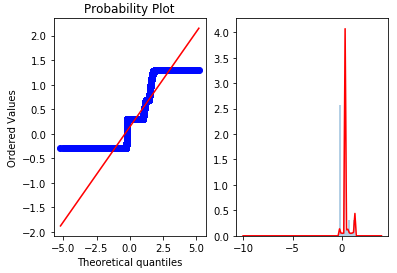} \\
  \includegraphics[width=3.79cm]{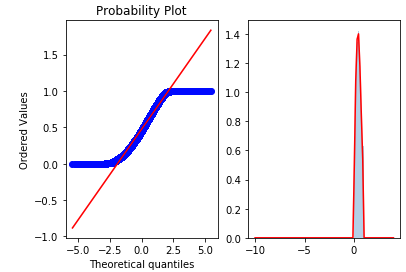} &
  \includegraphics[width=3.79cm]{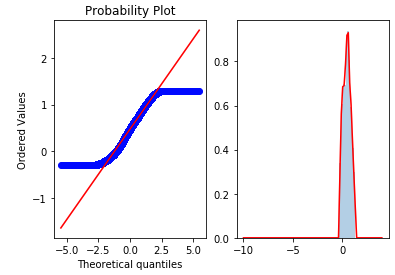} \\
  (a) & (b)
\end{tabular}
\caption{Probability density function for MNIST (a) and CIFAR10 (b)
before and after the adversarial perturbation}
\label{fig:distributions}
\end{figure}

Each model has been trained and its accuracy has been recorded on the
clean test data as well as on the adversarial one.
Moreover, we represent visually the distributions for each dataset and
how they change accordingly before and after the adversarial attacks
in Figure~\ref{fig:distributions}. Notice that in real world datasets
such as CIFAR10 the shift of the distribution is almost unnoticeable
which demonstrates the severity of adversarial attacks undermining the
security of neural networks. The first row contains the qq-plot along
with its density plot for MNIST before (a) and after (b) the
adversarial perturbation while the second row contains the same
information for CIFAR10.

We also demonstrate visually which are the most
vulnerable parts in regards to the top predictions that each neural
network has learned in order to differentiate two similar data points
belonging in the same category in
Figure~\ref{fig:misclass_autoenc}
for MNIST and Figure~\ref{fig:misclass_convnet_cifar},
Figure~\ref{fig:misclass_resnet_cifar} and
Figure~\ref{fig:misclass_autoenc_cifar} for CIFAR10.

Finally, we propose dictionary learning as a defense mechanism
against adversarial attacks. We demonstrate its ability on two
different adversarial attacks and record the results in
Table~\ref{tbl:fgsm} and Table~\ref{tbl:jsma}, which clearly shows
that, in overall, each model achieves higher classification accuracy
on the perturbed datasets after utilizing dictionary learning to
reconstruct the original data from the perturbed samples. One of the
advantages of dictionary learning is that it can operate regardless of
the presence of perturbed or noisy samples. This implies that the
dictionary $\mathbf{D}$ can be learned either from the extracted
patches of the perturbed samples or from the clean patches of the non
perturbed train data.  This could be useful in environments where we
do not know a priory the attack or the method used to generate the
perturbed data.  Another advantage of dictionary learning and sparse
coding is the fact that it can be embedded in any supervised learning
algorithm without any severe restrictions. In the particular case of
DNN the dictionary $\mathbf{D}$ can be easily learned from the weights
$\mathbf{W}$ of an auto-encoder model for instance.

Furthermore, the computational complexity of dictionary learning is
much lower compared to~\cite{wang2016:using-non-invertible-data} since
the dictionary can be learned during training time avoiding the
overhead in invoking a non invertible transformation during test
time. This means that whenever a prediction is required from the
model, a non invertible and computationally expensive transformation
has to be performed in advance.

\section{Experiments}
\subsection{Evaluation}
The experiments in this study utilized five DNN models that resemble
as close as possible real world application architectures composed of
multiple layers such as, dropout and batch normalization to avoid
over-fitting. We deliberately avoided DNN models composed of only
convolutional layers which seem to be more error prone to adversarial
attacks. The description of the hyper-parameters for each model is
provided in Table~\ref{tab:hyperparams}. Each model has been evaluated
on two different datasets perturbed using two different perturbation
techniques (FGSM) and (JSMA).  After the evaluation of dictionary
learning as defense mechanism against adversarial perturbations we
found that is able to withstand the attacks and provide
good results in terms of accuracy for each model on the reconstructed
datasets.
During training for consistency we utilized
Adam~\cite{kingma2014:adam-optimizer} as the optimizer for all the
models. We proceeded by perturbing the test set
$\mathbf{Y}\in\mathbb{R}^{M\times N}$ once for FGSM
$\mathbf{Y}^{\dagger}$, and once for JSMA $\mathbf{Y}^{\star}$, where
we tested each model equivalently on $\mathbf{Y},
\mathbf{Y}^{\dagger}, \mathbf{Y}^{\star}$, and recorded their
accuracies in Table~\ref{tbl:fgsm} and Table~\ref{tbl:jsma}. For each
image $\vec{y}\in \mathbf{Y}$ we extracted a set of overlapping
$8\times 8$ patches which were used to learn the dictionary
$\mathbf{D}$ and its coefficients $\vec{s}$ for each dataset as it is
shown below.

\begin{figure}
  \begin{tabular}{cc}
    \includegraphics[width=3.79cm]{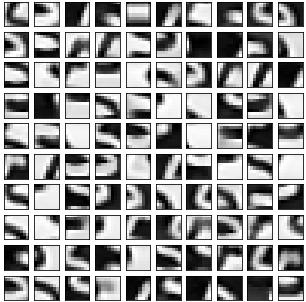} &
    \includegraphics[width=3.79cm]{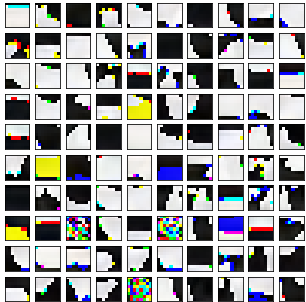}  \\
    (a) & (b)\\
    \multicolumn{2}{c}{Figure 4a: OMP dictionary, MNIST (a) \& CIFAR10 (b).}
  \end{tabular}
\end{figure}

Next we extracted patches from the noisy samples
$\mathbf{Y}^{\dagger}$ and $\mathbf{Y}^{\star}$, applying orthogonal
matching pursuit described in Algorithm 1 
reconstructing the original samples $\vec{y}\in\mathbf{Y}$. For each
sample we evaluated its reconstruction error utilizing mainly two
different metrics. The first one is referred to as peak to signal
noise ratio (PSNR) and is formulated as $PSNR =
10\log_{10}(\frac{{\max(\vec{y})}^{2}}{MSE})$, where $\max(\vec{y})$
refers to the maximum pixel value of the image $\vec{y}$ and
$\mbox{MSE}$ refers to the mean square error. The second one is
structural similarity index (SSIM) and is formulated according to
$SSIM(x, y) = \frac{(2\mu_{x}\mu_{y} + c_{1})(2\sigma_{xy} +
c_{2})}{(\mu_{x}^{2} + \mu_{y}^{2} + c_{1})(\sigma_{x}^{2} +
\sigma_{y}^{2} + c_{2})}$ where $x, y$ here represent windows of size
$N\times N$, $\mu$ represents the average of a window depending on the
subscript, $\sigma^{2}$ is the variance for each window and
$\sigma_{xy}$ is the co-variance between $x$ and $y$. Finally, $c_{1}$
and $c_{2}$ are constants in order to stabilize the division with weak
denominators.

\subsection{Results}
The results are summarized in Table~\ref{tbl:fgsm} and
Table~\ref{tbl:jsma} as well as in Figure 5 through Figure
9. Table~\ref{tbl:fgsm} describes the results for all five models by
evaluating their accuracy on MNIST and CIFAR10, on the actual test
data, on the perturbed one after the FGSM perturbation attack and
finally on the denoised samples recovered through denoising dictionary
learning. The choice of atoms was based on a heuristic selection and
it resulted in 38 atoms for MNIST and 2 atoms for CIFAR10. Although we
suspect that the overall results could be improved by utilizing
Bayesian hyper-parameter selection for the choice of atoms.
Table~\ref{tbl:jsma} equivalently describes the results for the actual
test data, perturbed under the JSMA perturbation attack and the
denoised one recovered through dictionary learning.

\begin{table}[ht]
\caption{Classifier accuracy against FGSM with noise intensity, $\epsilon = 0.3$}
\resizebox{\columnwidth}{!}{%
  \begin{tabular}{llcccccc}
    \toprule
    Dataset & & \multicolumn{3}{c}{MNIST} & \multicolumn{3}{c}{CIFAR10} \\
    \midrule
    Perturbed & & \ding{55} & \ding{51} & Denoised
                      & \ding{55} & \ding{51} & Denoised \\
    \midrule
    \multirow{5}{*}{Classifier} & MLP & 98.39\% & 12.80\% & 82\% &
    60\% & 11\% & 55.60\% \\
    \hhline{~-------}
     & ConvNet & 99.35\% & 79.90\% & 90.46\% & 77.90\% &
      15.06\% & 68.29\% \\
     \hhline{~-------}
     & AutoEnc & 99.34\% & 77.60\% & 89.37\% & 70.56\% &
      13.63\% & 67.76\% \\
     \hhline{~-------}
     & ResNet & 93.79\% & 1.95\% & 74.27\% & 76.11\% & 0.089\% &
       70.06\% \\
     \hhline{~-------}
     & HRNN & 98.90\% & 23.02\% & 82.52\% & 64\% & 15.75\% & 58.41\% \\
    \bottomrule
  \end{tabular} %
}
\label{tbl:fgsm}

\caption{Classifier accuracy against JSMA.}
\resizebox{\columnwidth}{!}{%
  \begin{tabular}{llcccccc}
    \toprule
    Dataset & & \multicolumn{3}{c}{MNIST} & \multicolumn{3}{c}{CIFAR10} \\
    \midrule
    Perturbed & & \ding{55} & \ding{51} & Denoised
                      & \ding{55} & \ding{51} & Denoised \\
    \midrule
    \multirow{5}{*}{Classifier} & MLP & 98.39\% & 53.02\% & 60\% &
    60\% & 52.77\% & 57.63\% \\
    \hhline{~-------}
     & ConvNet & 99.35\% & 79.90\% & 93.85\% & 77.90\% &
      14.59\% & 67.23\% \\
     \hhline{~-------}
     & AutoEnc & 99.34\% & 62.02\% & 91.47\% & 70.56\% &
      56.83\% & 64.76\% \\
     \hhline{~-------}
     & ResNet & 93.79\% & 38.16\% & 61.12\% & 76.11\% & 56.87\% &
       60.16\% \\
     \hhline{~-------}
     & HRNN & 98.90\% & 52.65\% & 63.21\% & 64\% & 56.43\% & 61.21\% \\
    \bottomrule
  \end{tabular} %
}
\label{tbl:jsma}
	\caption{Hyper-parameters for DNN models}
	\resizebox{\columnwidth}{!}{%
		\begin{tabular}{lccccc}
			& \bf{MLP} & \bf{ConvNet} & \bf{AutoEncoder} & \bf{ResNet} & \bf{HRNN} \\
			\hhline{~-----}
			& Dropout 0.5 & Dropout 0.5 & Conv2D: filters=16, kernel=3 &
			Conv2D &
			LSTM: units=256\\
			\hhline{~-----}
			& BatchNorm & 2 $\times$ Conv2D: filters=32, kernel=3 & MaxPool: size=2
			& BatchNorm & TimeDistributed\\
			\hhline{~-----}
			& Dense 784 & MaxPool: size=2 & Conv2D: filters=32, kernel=3 & ReLU
			& LSTM: units=256\\
			\hhline{~-----}
			& ReLU & Dropout 0.25 & MaxPool: size=2 & Basic Block & Reshape: 16$\times$16 \\
			\hhline{~-----}
			& Dropout 0.2 & 2 $\times$ Conv2D: filters=64, kernel=3 & Conv2D:
			filters=32,
			size=3 &
			3$\times$Residual Block
			&
			LSTM: units=256 \\
			\hhline{~-----}
			& BatchNorm &  MaxPool: size=2 & MaxPool: size=2 & Dropout 0.25 & Dropout 0.25 \\
			\hhline{~-----}
			& Dense 256 & Conv2D: filters=128, kernel=3 & Conv2D: filters=64,
			kernel=3 & BatchNorm
			& Dense 10\\
			\hhline{~-----}
			& ReLU &  Conv2D: filters=256, kernel=3 & Conv2D: filters=128,
			kernel=3 & ReLU & Softmax \\
			\hhline{~-----}
			& Dense 10 & Dropout 0.25 & UpSample: size=2 &
			GlobalAveragePooling & \\
			\hhline{~-----}
			& Softmax & Dense 512 & Conv2D: filters=32,
			kernel=3 & Dense 10 & \\
			\hhline{~-----}
			&  & Dropout 0.5 & UpSample: size=2 & Softmax & \\
			\hhline{~-----}
			&  & Dense 10 & Conv2D: filters=3, kernel=3 &  & \\
			\hhline{~-----}
			&  & Softmax & Conv2D: filters=1, kernel=5 &  & \\
			\hhline{~-----}
			&  &  & Dense 10 & & \\
			\hhline{~-----}
			&  &  & Softmax & & \\
			\hhline{~-----}
		\end{tabular}%
	}
\label{tab:hyperparams}
\end{table}

\begin{figure}[ht]
		\begin{tabular}{c}
			\includegraphics[width=8cm]{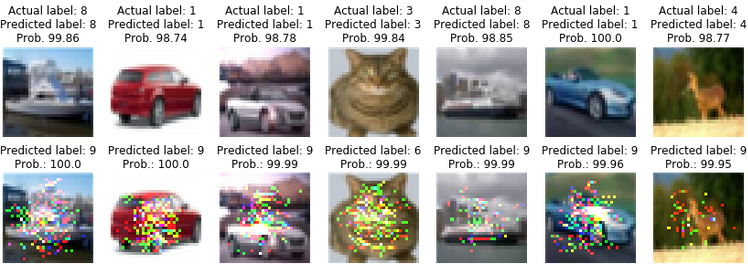} \\
			\includegraphics[width=8cm]{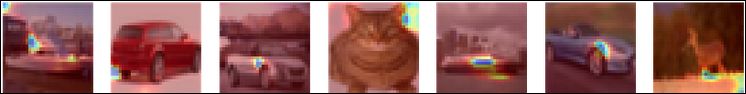}
		\end{tabular} %
\caption{Misclasification for ResNet CIFAR10.}
\label{fig:misclass_resnet_cifar}
\end{figure}

\begin{figure}[ht]
		\begin{tabular}{c}
			\includegraphics[width=8cm]{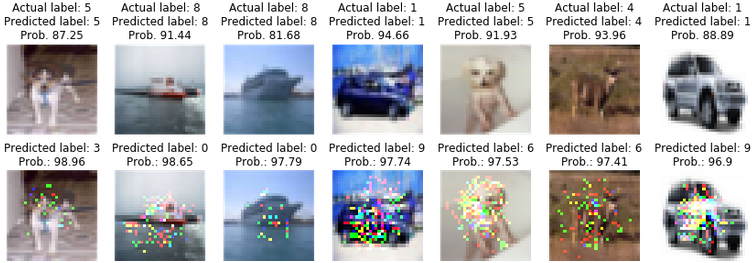}  \\
			\includegraphics[width=8cm]{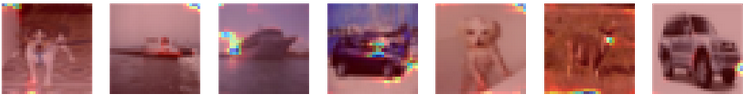}
		\end{tabular} %
\caption{Misclasification for AutoEenc. on CIFAR10.}
\label{fig:misclass_autoenc_cifar}
\end{figure}

\begin{figure} [ht]
\begin{tabular}{c}
	\includegraphics[width=8cm]{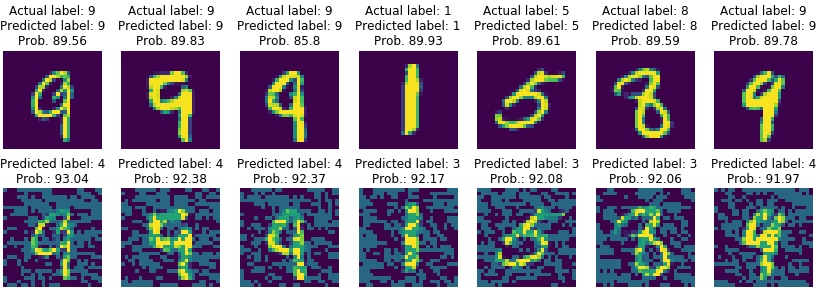} \\
	\includegraphics[width=8cm]{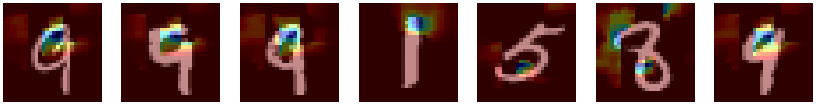}
\end{tabular} %
\caption{Misclasification for AutoEenc. on MNIST.}
\label{fig:misclass_autoenc}
\end{figure}

\begin{figure}[ht]
  \begin{tabular}{cc}
    \includegraphics[width=3.74cm]{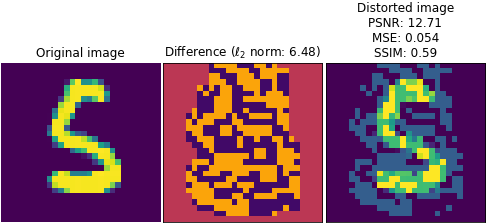} &
    \includegraphics[width=3.74cm]{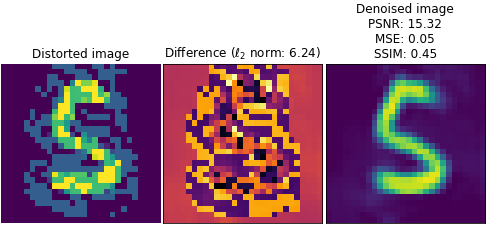} \\
    \multicolumn{2}{c}{(a)} \\
	\includegraphics[width=3.74cm]{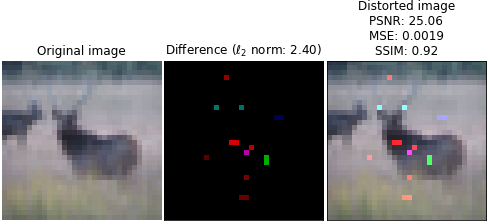} &
    \includegraphics[width=3.74cm]{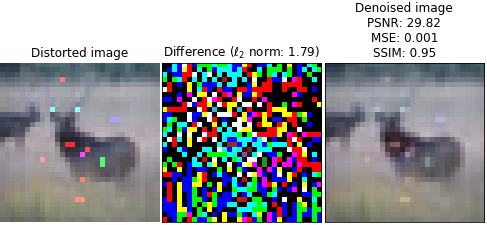} \\
    \multicolumn{2}{c}{(b)}
  \end{tabular} %
\caption{Denoising on MNIST, FGSM (a) \& CIFAR10, JSMA (b).}
\label{fig:adv_example}
\end{figure}

\begin{figure}[!ht]
\begin{tabular}{c}
  \includegraphics[width=8cm]{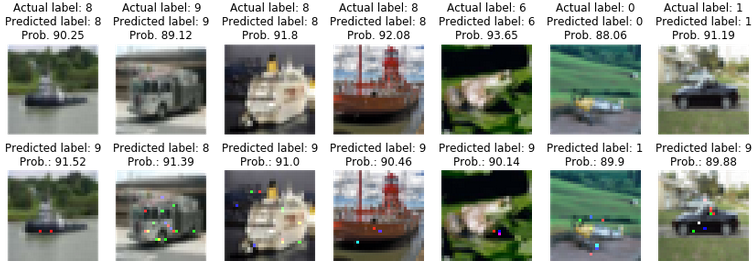} \\
  \includegraphics[width=8cm]{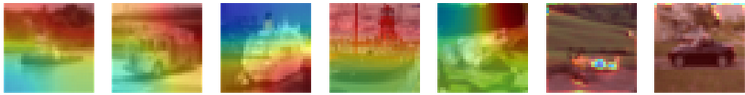}
\end{tabular} %
\caption{Misclasification for ConvNet CIFAR10.}
\label{fig:misclass_convnet_cifar}
\end{figure}

As it is evident all five models achieve higher accuracy on the
recovered samples compared to the perturbed version.
We present the top classifications for the convolutional and residual
network model along with their misclasification on the perturbed
data under the FGSM attack on MNIST, as well as, their class
activation maps which describe the sensitivity of the classifier on
different parts of the input.
In Figure~\ref{fig:misclass_convnet_cifar} and
Figure~\ref{fig:misclass_resnet_cifar} we present the top
misclasification for the convolutional and residual network along
with the activation maps for CIFAR10 perturbed under the JSMA attack.
In Figure~\ref{fig:misclass_autoenc} we show the top classifications
for the auto-encoder model along with their activation maps under
the FGSM attack on MNIST. Equivalently in
Figure~\ref{fig:misclass_autoenc_cifar} we show the top
misclasification for the auto-encoder on
CIFAR10 under the JSMA attack. As you might have noticed the
multilayer perceptron does not have feature maps similar to a
convolutional network therefore it is impossible to derive the class
activation maps. Similarly the same holds true for the hierarchical
recurrent model. We noticed that the results from the residual network
were not as resistant as we would have expected due to its skip
connections. What we can infer from
Figure~\ref{fig:misclass_resnet_cifar} is that models who have the
ability to focus on very small details of the overall image seem to be
more susceptible to adversarial perturbations. In
Figure~\ref{fig:adv_example} we demonstrate an example of a
reconstructed image equivalently for each dataset and perturbation
attack.

\section{Conclusion}
In this article, a defense mechanism is proposed against adversarial
perturbations based on dictionary learning sparse representations for
gray scale (MNIST) and color images (CIFAR10). The method has been
evaluated against five modern deep neural network architectures which
compose the building blocks for the majority of recent neural network
architectures. The choice of dictionary learning is based solely on
its properties. The resulted dictionary is a redundant, over-complete
basis, and it provides a more efficient representation than a normal
basis. It is robust against noise, it has more flexibility for
matching patterns in the data, and it allows a more compact
representation. Future directions include the extension and comparison
of the current work with deep denoising models such as gated Markov random fields and deep Boltzmann machines on the ImageNet
dataset.

 \begin{figure}[!ht]
 \begin{tabular}{c}
   \includegraphics[width=8cm]{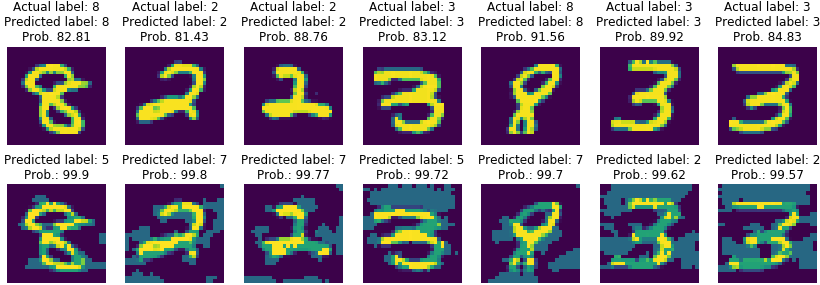} \\
   \includegraphics[width=8cm]{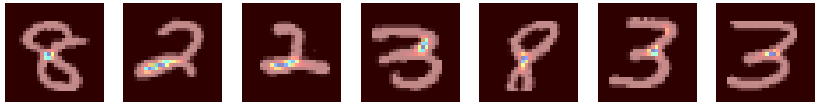}
 \end{tabular} %
\caption{Misclassification for ResNet on MNIST.}
\label{fig:res-top-predictions-mnist}
\end{figure}

\bibliographystyle{aaai}
\bibliography{biblio}
\end{document}